\documentclass[conference]{IEEEtran}

\usepackage{microtype}
\usepackage{graphicx}
\usepackage{subfigure}
\usepackage{booktabs}

\usepackage{siunitx}
\usepackage{soul,color}
\usepackage{amsmath}
\usepackage{amssymb}

\usepackage{hyperref}

\pdfoutput=1

\begin{document}

\title{Pseudo-Recursal: Solving the Catastrophic Forgetting Problem in Deep Neural Networks}

\author{\IEEEauthorblockN{Craig Atkinson}
\IEEEauthorblockA{Department of Computer Science\\
University of Otago\\
Dunedin, New Zealand\\
Email: atkcr398@student.otago.ac.nz}
\and
\IEEEauthorblockN{Brendan McCane}
\IEEEauthorblockA{Department of Computer Science\\
University of Otago\\
Dunedin, New Zealand\\
Email: mccane@cs.otago.ac.nz}
\and
\IEEEauthorblockN{Lech Szymanski}
\IEEEauthorblockA{Department of Computer Science\\
University of Otago\\
Dunedin, New Zealand\\
Email: lechszym@cs.otago.ac.nz}
\and
\IEEEauthorblockN{Anthony Robins}
\IEEEauthorblockA{Department of Computer Science\\
University of Otago\\
Dunedin, New Zealand\\
Email: anthony@cs.otago.ac.nz}
}

\maketitle

\begin{abstract}
In general, neural networks are not currently capable of learning tasks in a sequential fashion. When a novel, unrelated task is learnt by a neural network, it substantially forgets how to solve previously learnt tasks. One of the original solutions to this problem is pseudo-rehearsal, which involves learning the new task while rehearsing generated items representative of the previous task/s. This is very effective for simple tasks. However, pseudo-rehearsal has not yet been successfully applied to very complex tasks because in these tasks it is difficult to generate representative items. We accomplish pseudo-rehearsal by using a Generative Adversarial Network to generate items so that our deep network can learn to sequentially classify the CIFAR-10, SVHN and MNIST datasets. After training on all tasks, our network loses only 1.67\% absolute accuracy on CIFAR-10 and gains 0.24\% absolute accuracy on SVHN. Our model's performance is a substantial improvement compared to the current state of the art solution.
\end{abstract}

\section{Preface}
We have released this article in good faith and have recently been made aware of a comparable method that predates ours~\cite{shin2017continual}. For a different perspective on how pseudo-rehearsal with Generative Adversarial Networks can be used to solve the catastrophic forgetting problem, please refer to their article.

\section{Introduction}
Deep Neural Networks (DNNs) are currently the state of the art solution to many machine learning problems. However, they cannot be trained on a new task while retaining knowledge of previously trained tasks. This is known as Catastrophic Forgetting (CF)~\cite{mccloskey1989catastrophic} and is a problem that needs to be solved for continuously learning artificial agents - so called lifelong learners.

Solutions to CF can be aligned with the biological learning notions of synaptic stability and plasticity~\cite{abraham2005memory}. The synaptic stability hypothesis states that memory is retained by fixing the weights between the units that encode it. The synaptic plasticity hypothesis states that weights between units can change as long as memory is retained such that the output units still recreate the correct pattern of activity.

Recent focus on overcoming CF has introduced Elastic Weight Consolidation (EWC)~\cite{kirkpatrick2017overcoming}. In EWC, the loss function is augmented by the importance of weights to previously learnt tasks. This measure encourages weights that are very important to the previous task to retain similar values, whereas less important weights can be more significantly altered to learn the new task. EWC is aligned with the synaptic stability hypothesis because it tries to reduce significant changes to important weights which would impair the functionality of the network and thus, its performance on the previous task. However, constraining each of the important neurons in the network to retain similar weights has a number of disadvantages. Predominantly, if a group of neurons' function in the network can be compressed into a smaller group of neurons (to make room for new information), EWC will not find this solution. Another disadvantage with EWC is that two task specific parameters were needed per unit to allow this method to work on challenging problems such as playing Atari 2600 games.

Pseudo-rehearsal~\cite{robins1995catastrophic} has been proposed as a solution to CF and thus, life long learning in neural networks. When the network needs to be trained on a new task, pseudo-rehearsal protects prior learning by generating samples of the network's behaviour which captures the structure of the original task/s. This is done by randomly generating input samples and assigning their target outputs by passing them through the network. These input-output pairs can then be rehearsed while learning items from the new task. Pseudo-rehearsal constrains the changes to the network's function so that it remains approximately the same for the input space of previously learnt tasks while the function changes for the input space of the new task. Pseudo-rehearsal is aligned with the synaptic plasticity hypothesis because the weights of individual neurons encoding the previous task can change as long as the network's overall functionality remains consistent.

\section{Pseudo-Recursal}
Let's begin with formalising pseudo-rehearsal. Let $\mathbf{x} \in \mathbb{R}^d$ and a neural network of a chosen architecture with its function given as $h_w: \mathbb{R}^d \mapsto \mathbb{R}^k$, where $w$ is the set of all trainable parameters. Let $y_t$ be a one-hot encoded vector indicating the assignment of one class in task $t$, with all zero vectors for other classes across all tasks $t = 1, ..., T$.

Given the set of parameters $w_t$, which minimises some loss function $\mathcal{L}\left(h_{w_t}(\mathbf{x}_t), \mathbf{y}_t\right)$ for task $t$, we want to next train the network on task $t+1$ and find an optimal set of weights $w_{t+1}$ that minimises loss $\mathcal{L}\left(h_{w_{t+1}}(\mathbf{x}_{t+1}), \mathbf{y}_{t+1}\right)$ while still performing well on the previous task, i.e. $h_{w_{t+1}}(\mathbf{x}_t) \approx h_{w_t}(\mathbf{x}_t)$. Now, if we were to rehearse the previous tasks while learning task $t+1$ we would need to minimise the following:

\begin{equation}
J_{r} = \sum\limits_{i = 1}^{t + 1} \mathcal{L}\left(h_{w_{t+1}}(\mathbf{x}_{i}), \mathbf{y}_{i}\right)
\end{equation}

It's problematic, because it needs all the data from the previous tasks.

In pseudo-rehearsal, the same can be achieved by minimising:

\begin{equation}
J_{p} = \mathcal{L}(h_{w_{t+1}}\left(\mathbf{x}_{t+1}), \mathbf{y}_{t+1}\right) + \sum\limits_{i = 1}^{t} \mathcal{L}\left(h_{w_{t+1}}(\mathbf{\widetilde{x}}_i), \mathbf{\widetilde{y}}_i\right),
\end{equation}

where $\mathbf{\widetilde{x}}_{i}$ is a pseudo-vector generated randomly and $\mathbf{\widetilde{y}}_{i} = h_{w_t}(\mathbf{\widetilde{x}}_{i})$ is the output of the network at state $w_t$, before learning task $t + 1$.

The problem is that for image datasets, random vector $\mathbf{\widetilde{x}}_i$ is not representative of the input $\mathbf{x_i}$.

\subsection{Generating Representative Pseudo-Images}

\begin{figure}[ht]
\vskip 0.2in
\begin{center}
\centerline{\includegraphics[clip,trim={3.4cm 9.6cm 15.6cm 3.1cm},width=\columnwidth]{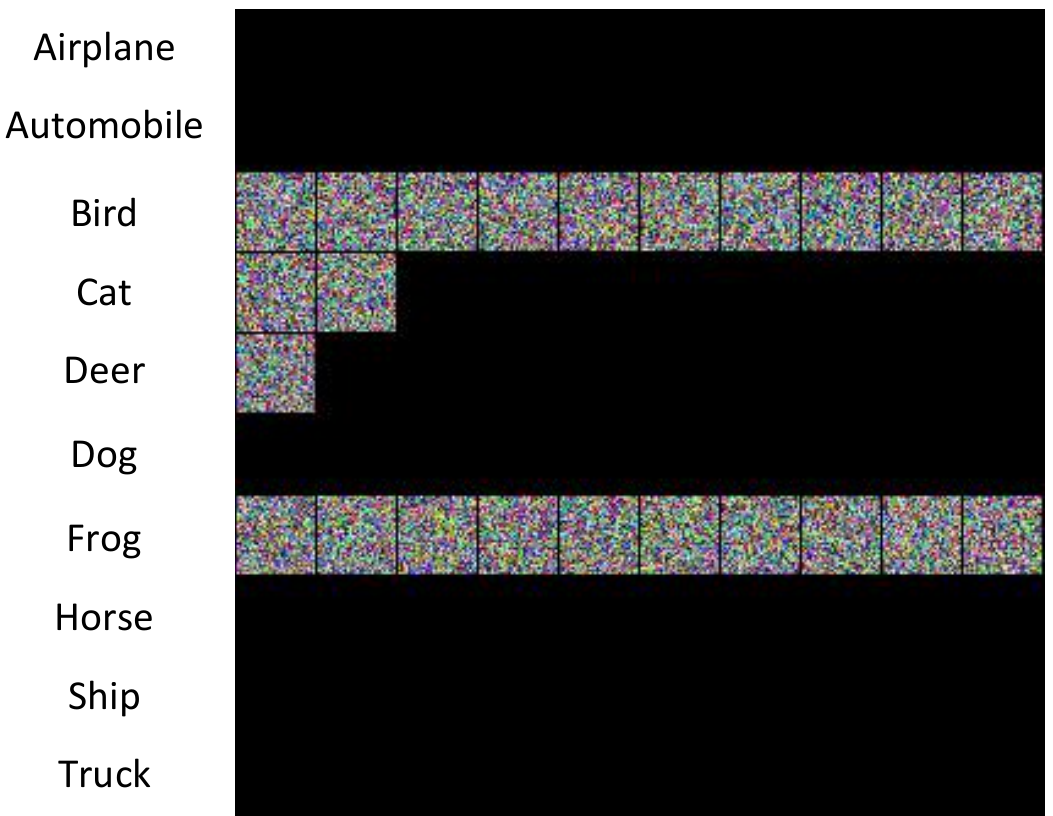}}
\caption{Pseudo-images generated by a uniform distribution $[0, 255]$. These images are labelled by our classification network. Images are black when no instances of that class occurred after 2048 generations.}
\label{tiled-images-static}
\end{center}
\vskip -0.2in
\end{figure}

\begin{figure}[ht]
\vskip 0.2in
\begin{center}
\centerline{\includegraphics[clip,trim={0.5cm 20.4cm 6.4cm 0.5cm},width=\columnwidth]{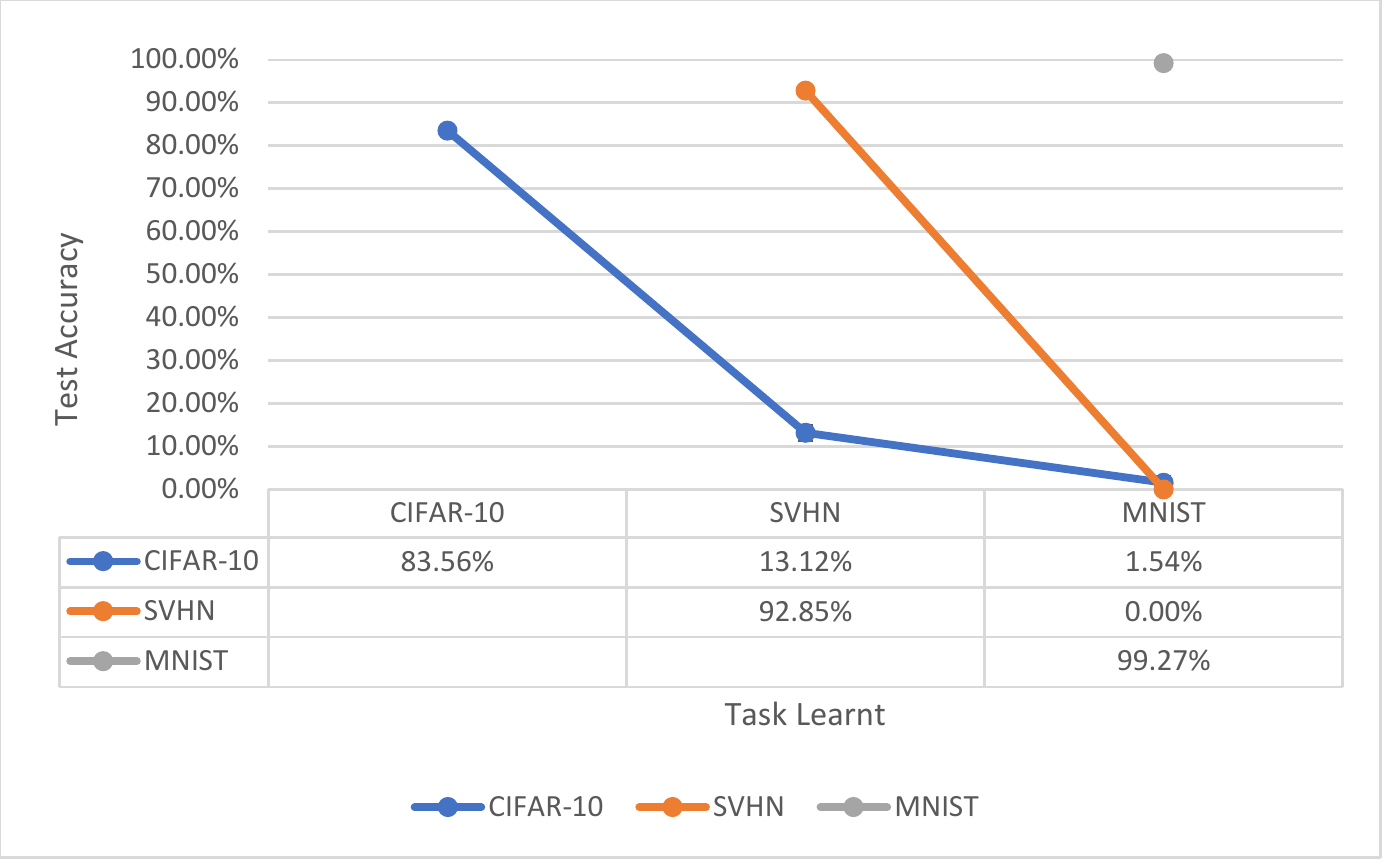}}
\caption{Average accuracy of a classification network when using pseudo-rehearsal with pseudo-images generated by a uniform distribution $[0, 255]$. The x-axis represents the task that has just been learnt and the lines represent the network's test accuracy on the various tasks trained so far. Error bars represent the standard deviation of each data point across the 3 trials. Non-visible error bars have smaller standard deviations than their data point.}
\label{results-rand}
\end{center}
\vskip -0.2in
\end{figure}

In~\cite{robins1995catastrophic}, pseudo-rehearsal was achieved by generating pseudo-items using a random distribution. However, in deep learning, the problems are much harder and thus, pseudo-vectors generated purely at random are not likely to be good representations of the training data. This is particularly obvious for images because generating pseudo-images with a uniform distribution produces static images which do not represent natural images (see Figure~\ref{tiled-images-static}). Furthermore, these static images poorly represent the distribution of classes as, after 2048 generations, the network believed almost all static images were either birds or frogs. When these static images are used in pseudo-rehearsal, the network retains little knowledge of its previously learnt tasks (see Figure~\ref{results-rand}).

The Generative Adversarial Network (GAN)~\cite{goodfellow2014generative} is a neural network model which uses unsupervised learning to generate random images which are representative of the input dataset. This is achieved by creating two network models; a discriminative model and a generative model. The goal of the discriminative model is to identify whether an input image is a real image or a generated image, whereas the goal of the generative model is to create images which fool the discriminator. This results in the generator learning to create images that represent the input images.

\begin{figure*}[ht]
\vskip 0.2in
\begin{center}
\centerline{\includegraphics[clip,trim={1.2cm 9.0cm 1.1cm 2.8cm},width=\textwidth]{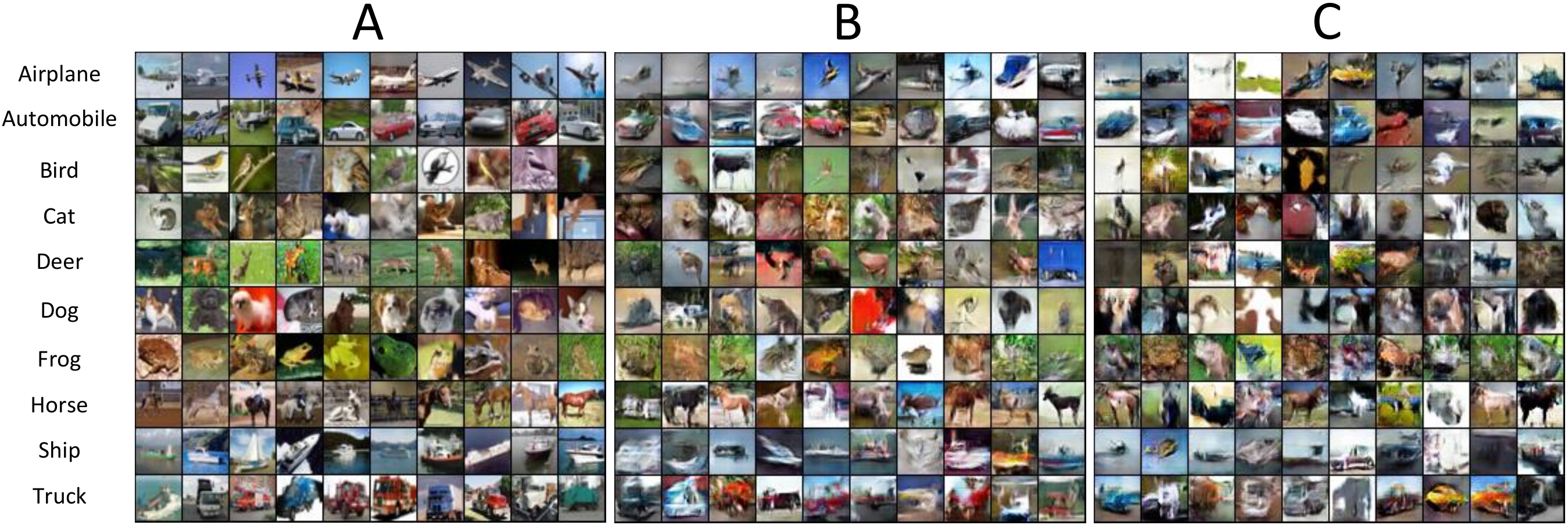}}
\caption{\textbf{A}: Real CIFAR-10 images. \textbf{B}: Pseudo-images generated by a GAN trained on CIFAR-10 images. \textbf{C}: Pseudo-images generated by a GAN trained on images from CIFAR-10, followed by images from SVHN along with pseudo-images representing CIFAR-10. These images are labelled by our classification network. The GAN from \textbf{C} generates images representing both SVHN and CIFAR-10, however only images representing (labelled by the classifier as) CIFAR-10 are included in this figure.}
\label{tiled-images-cifar10}
\end{center}
\vskip -0.2in
\end{figure*}

To confirm that a GAN can be used to generate pseudo-images that look similar to real images, we applied a Deep Convolutional Generative Adversarial Network (DCGAN)~\cite{radford2015unsupervised} to the CIFAR-10 dataset\footnote{A few adaptations were made to this algorithm. These are described in Section~\ref{network-arch}.}. Figure~\ref{tiled-images-cifar10} (B) illustrates that the generated images look similar to real CIFAR-10 images from a distance, although differences emerge on close inspection. Nevertheless, the generated images still contain class specific features which the network can learn to retain.

\subsection{Learning Process}
Our GAN is trained on images from a previous task so that it can be used to generate pseudo-images which are representative of that task, without storing the actual dataset. This means that when the classification network is trained on the first task (T1), the GAN should also be trained on T1's images. Then, when the next task (T2) is learnt, the GAN can be used to generate pseudo-images (which have their target labels assigned by the previous classification network) which are rehearsed along with T2.

The main complication is when a third task (T3) is introduced. To retain knowledge of both T1 and T2 our pseudo-items must now be representative of both tasks. A simple solution is to train a second GAN to generate images from T2. This is effective, but requires a new allocation of memory for every task. A more elegant solution is to do pseudo-rehearsal on the GAN as well. This allows the GAN to produce pseudo-items from both T1 and T2 without requiring extra memory per task. Pseudo-rehearsal on the GAN model can easily be achieved by generating pseudo-images from the GAN and mixing them with the current task's images. This procedure can be repeated recursively, every time a new task is present and thus, we term our process pseudo-recursal. 

\begin{figure*}[ht!]
\vskip 0.2in
\begin{center}
\centerline{\includegraphics[clip,trim={3.4cm 9.0cm 7.2cm 2.8cm},width=\textwidth]{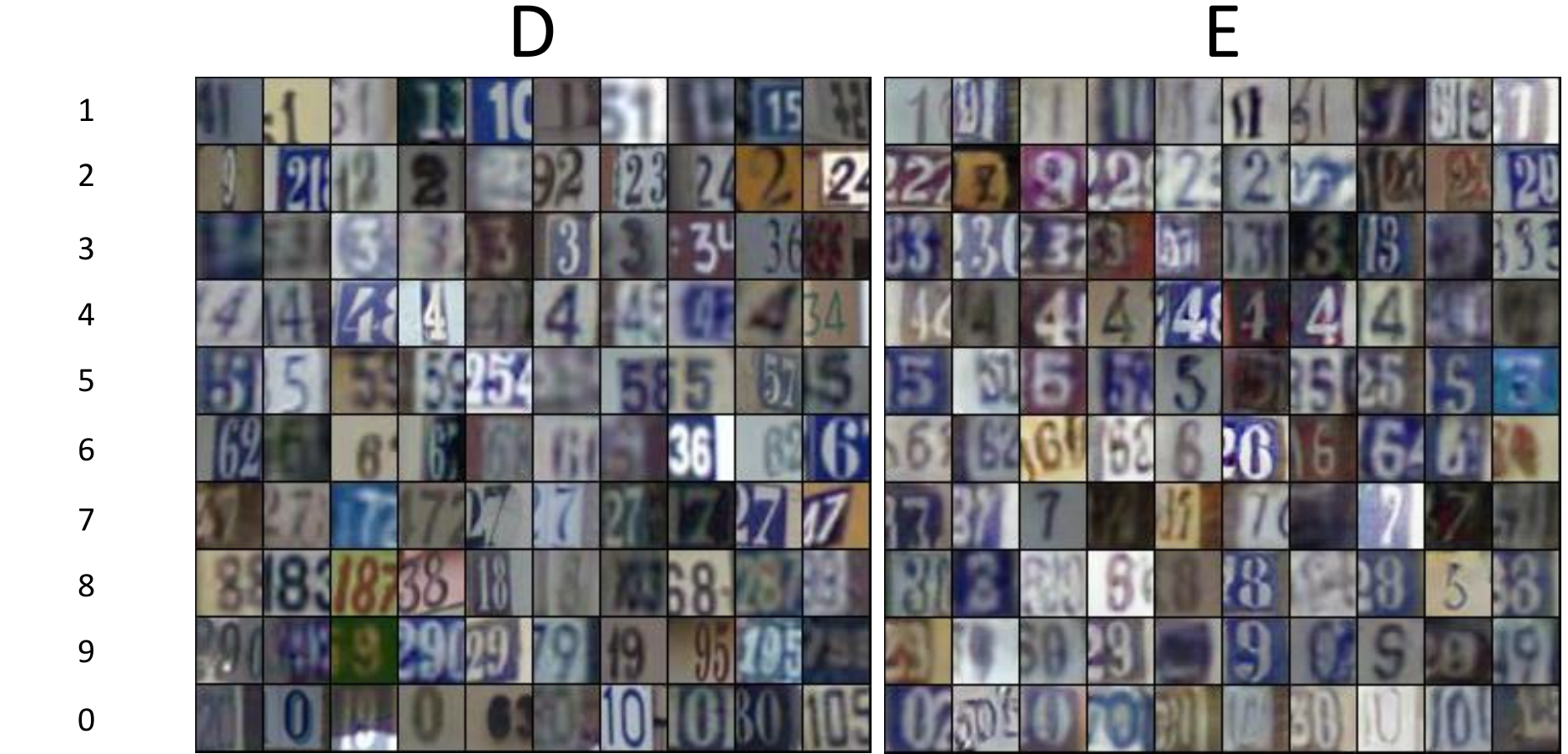}}
\caption{\textbf{D}: Real SVHN images. \textbf{E}: Pseudo-images generated by a GAN trained on images from CIFAR-10, followed by images from SVHN along with pseudo-images representing CIFAR-10. These images are labelled by our classification network. The GAN from \textbf{E} generates images representing both SVHN and CIFAR-10, however only images representing (labelled by the classifier as) SVHN are included in this figure.}
\label{tiled-images-svhn}
\end{center}
\vskip -0.2in
\end{figure*}

To confirm that pseudo-rehearsal can be used on the GAN to still generate pseudo-images that look similar to real images, we train our GAN on CIFAR-10 and then SVHN while rehearsing generated images that represent CIFAR-10. Figure~\ref{tiled-images-cifar10} and Figure~\ref{tiled-images-svhn} illustrates that using pseudo-rehearsal on the GAN causes it to generate images that represent the recent task (SVHN) and the previously trained task/s (CIFAR-10). Furthermore, in the case of SVHN, the pseudo-images do not appear to be noticeably different from real SVHN images.

Originally, pseudo-rehearsal was applied to simple tasks where the pseudo-items generated by a random distribution covered the whole input space. This means that the pseudo-items represent the network's function over the whole input space such that changes that are made to accommodate the new task are as local as possible to the input space of the new task~\cite{robins1998local}. However, we are using a GAN so that we only generate pseudo-items near actual previous inputs in a much larger / sparser space and thus, the network retains its mapping only near previous inputs, and in other parts of the space, the network is free to vary.

In summary, we achieve continuous learning by applying pseudo-rehearsal to both a classification model and a GAN model. This allows our DNN to overcome the CF problem without requiring extra memory when a new task is presented. Furthermore, our pseudo-recursal technique does not apply any hard constraints on the function of intermediate neurons nor the function of the network in irrelevant areas of the input space.

\section{Related Work}
Our aim is to develop a general purpose algorithm for overcoming the CF problem in sequential task learning. We believe that for this algorithm to be general purpose and scalable to a large number of tasks, the following criteria should be met:
\begin{enumerate}
	\item It should be able to be applied to the entirety of a DNN without any layers being pre-trained on similar tasks.
	\item It should not require the memorisation of any past tasks' items.
	\item It should not grow in memory requirements with each new task.
	\item The function of intermediate neurons should not be constrained so that they adhere to the synaptic plasticity hypothesis.
\end{enumerate}

Incremental learning is a field of research where a neural network is required to learn a task's classes sequentially. In incremental learning, CF impairs networks' ability to remember previously trained classes. Our paper focuses on sequentially learning different tasks. These tasks are generally dissimilar and do not necessarily have to have the same modality (such as visual or auditory). Although these two learning problems differ, they have many similarities and therefore, methods from one field are often applicable to the other.

Recently, pseudo-rehearsal has been applied to incremental learning models. In~\cite{mellado2017pseudorehearsal}, pseudo-rehearsal was used to sequentially train a convolutional neural network to classify the first 5 digits of MNIST, followed by the remaining 5 digits. This was achieved by using a recurrent neural network to generate random images representative of the first 5 MNIST digits. These images could then be rehearsed while learning the later 5 digits. This work has a number of major limitations. Their generative model requires each pixel's mean value and standard deviation to be stored for each of the classes learnt. This results in the memory requirements of the model scaling with the number of classes learnt. Furthermore, they must check each of the generated images and discard any that the network is not 95\% confident in belonging to a learnt class. Finally, generating random images that are representative of MNIST is a trivial task because all images in a class are very similar to one another, which is not the case for more difficult tasks such as CIFAR-10 and to a lesser extent SVHN.

In~\cite{kemker2017fearnet}, pseudo-rehearsal was also applied to incremental learning in a model called FearNet. Their model has 3 components: short-term memory system, long-term memory system and a system that determines whether the short-term or long-term system should be used for classifying an item. This model stores recent items in the short-term system and pseudo-rehearsal is periodically used to train the long-term system on these items, without forgetting its previously learnt classes. The long-term memory system is implemented as an autoencoder (encoder-decoder) where the output of the encoder is passed through a final classification layer. The decoder is used to generate pseudo-items, however it also requires the storage of the mean and covariance matrix of the encoder's representation of each of the previous classes.

\begin{table*}[t]
\caption{Hyper-parameters for the classification model.}
\label{hyper-params}
\vskip 0.15in
\begin{center}
\begin{small}
\begin{tabular}{c c lc}
\toprule
Parameter & Value & Description\\
\midrule
$initial\_lr$ & \num{1e-3} & Learning rate used when the network is training on only the first task's dataset.\\
$later\_lr$ & \num{1e-4} & Learning rate used when the network is being trained on any later task.\\
$minibatch\_size$ & 512 & The number of items trained from during each mini-batch.\\
$patience$ & 10 & Training is stopped when the network has not improved in its validation error for this number of epochs.\\
$\beta_1$ & 0.9 & First moment decay rate for the Adam optimiser.\\
$\beta_2$ & 0.999 & Second moment decay rate for the Adam optimiser.\\
$\epsilon$ & \num{1e-8} & Epsilon value for the Adam optimiser.\\
$p\_train\_size$ & 37,500 & Number of pseudo-items in the training portion of the pseudo-dataset.\\
$p\_valid\_size$ & 12,500 & Number of pseudo-items in the validation portion of the pseudo-dataset.\\
\bottomrule
\end{tabular}
\end{small}
\end{center}
\vskip -0.1in
\end{table*}

In~\cite{kemker2017fearnet}, authors train FearNet's long-term memory store with a process called intrinsic replay~\cite{draelos2016neurogenesis}. This process is similar to our recursive training method, where pseudo-items are generated by the decoder and then rehearsed with the new items. Our generative model is isolated from the classification network, whereas FearNet combines these models. This means that when the long-term system is being trained, the classification loss and reconstruction loss are minimised concurrently. A further difference is that the reconstruction loss is minimised across each layer of the autoencoder. This limitation means that the function of neurons in intermediate layers of the network are being constrained.

FearNet was shown to counteract CF on supposedly complex tasks such as CIFAR-100. However, FearNet was never trained on the CIFAR-100's raw images but rather the output of the first 49 weight layers (including the mean pooling layer) in ResNet-50~\cite{he2016deep} that had been pre-trained on ImageNet. This means that the classification is done by a much smaller multi-layer perceptron and the majority of the work has been pre-trained into the ResNet architecture. Subsequently, the autoencoder is not learning to reproduce CIFAR-100 images but rather the items' output from the ResNet architecture. Although this is a more difficult task than training on MNIST, this method is incomparable to training a convolutional neural network to classify the task from raw input.

A recent review of methods for overcoming CF concluded that current algorithms do not solve CF~\cite{kemker2017measuring}. They also found that EWC performed the best for learning multiple tasks and thus, we compare pseudo-recursal to EWC. The main contributions of our paper are; we show that pseudo-recursal can be used to overcome the CF problem in DNNs, sequentially learning CIFAR-10, SVHN and MNIST and we demonstrate that pseudo-rehearsal can be applied recursively to a separate classification and generative model. This architecture also satisfies the previously mentioned criteria. Although we limit ourselves to image classification in this paper, our techniques are applicable to other problems.

\section{Method}
\subsection{Datasets}
In our experiments we train our classifier model sequentially on the CIFAR-10, SVHN and MNIST datasets\footnote{Other variations of this order were also tested and similar results were found.}. These datasets have been chosen because they all comprise similar sized images, the same number of classes and a range of similarities and differences between the datasets' tasks. CIFAR-10 contains animals and types of transport which is dissimilar to SVHN and MNIST which both contain the digits 0-9. All datasets are divided so that there are 37,500 training, 12,500 validation and 10,000 testing items.

For the classification network, all the tasks' and pseudo-datasets' validation and test images are center cropped to $24 \times 24$ and then standardised. For the training images, distortions are applied every epoch by randomly cropping the $32 \times 32$ images down to $24 \times 24$, flipping images left or right (only for CIFAR-10), adjusting brightness between -63 and 63, adjusting contrast between 0.2 and 1.8 and then standardising the images.

\subsection{Network Architecture} \label{network-arch}
The classification network we use is based on \cite{krizhevsky2012imagenet}. Our network passes the input through two convolutional layers, max-pooling layer, two more convolutional layers, max-pooling layer and then 3 fully-connected layers. Convolutional layers had 128, 128, 256 and 256 filters respectively and applied these $3 \times 3$ filters with a stride of 1. Max-pooling layers applied a $3 \times 3$ window with a stride of 2. The fully-connected layers had 512, 384 and 30 units respectively, with the final 30 unit layer being the softmax layer. All layers except the softmax layer apply the ReLU activation function.

The GAN used in this paper is identical to DCGAN~\cite{radford2015unsupervised}. The only exceptions are the use of pseudo-rehearsal with a dataset size of 50,000 items and a mini-batch discrimination layer (see~\cite{salimans2016improved}). The mini-batch discrimination layer reduces the training time needed for the generator to produce visually appealing images and helps stop the network from converging at a point where it only outputs the same image.

\subsection{Training and Evaluation}

The classifier is trained using the hyper-parameters specified in Table~\ref{hyper-params}. When the first task is being trained, all of the mini-batch's training examples come from the task's dataset. However, for later tasks, half of the examples come from the task's dataset and the remaining are from the pseudo-dataset.

The validation error is recorded after each epoch on both the current task's dataset and the pseudo-dataset (if one exists). After training is completed, the network weights at the epoch with the lowest validation loss are reloaded into the network and it is evaluated on real test items from the current task and all previously learnt tasks.

\subsection{Experimental Conditions}

Each experimental condition underwent 3 trials and results were averaged. The experimental conditions are as follows:
\begin{itemize}
	\item $std$: Learns the datasets individually in a sequence. This is the lower bound on performance (CF).
	\item $reh$: Learns the datasets sequentially, while still rehearsing all of the real items from previously learnt datasets. This is the upper bound on performance, which the network cannot improve.
	\item $pseudo\_rec$: Learns the datasets sequentially, while rehearsing pseudo-items representative of the previously learnt datasets.
	\item $ewc$: Learns the datasets sequentially, while retaining past knowledge with EWC and using task specific weights\footnote{The network is correctly told which task it is classifying so that the correct task specific weights are always applied. This gives EWC the best possible chance at outperforming pseudo-recursal.}. EWC uses a $\lambda$ parameter which we set to 270 after doing a random search between $[0, 1000)$ with 20 trials.
	\item $ewc\_c10$: When EWC was proposed, it was shown on tasks that shared their output units, for example, in reinforcement learning of Atari 2600 games, the buttons on the controller were consistent across games (with the exception that a couple were not used for particular games). In our classification problem this does not make sense, as if the output neurons were shared between tasks, a separate network would be required to determine the current task. However, to determine whether EWC is more effective in this scenario, we create a further condition which assumes the current task is known and thus, the 10 output neurons are shared between tasks. EWC's $\lambda$ parameter is set to 267 after doing a random search between $[0, 1000)$ with 20 trials.
	\item $rote\_learn$: To confirm that using a GAN to generate pseudo-items is more effective than using the same allocation of memory to rote learn a subset of the previous tasks' items, we tested our model's retention when rehearsal was applied to the subset of items. The number of free parameters in the generative model is approximately 4.5m and thus, 1,500 images (and their true labels) were randomly selected to be memorised from past tasks. This condition learns the datasets sequentially, while still rehearsing the memorised items. The images are split between the training and validation sets in the same 3:1 ratio as all other datasets and distortions are also applied to the training images every epoch.
\end{itemize}

\section{Results}

\begin{figure*}[ht!]
\vskip 0.2in
\begin{center}
\centerline{\includegraphics[clip,trim={.5cm 12.7cm 1.8cm .5cm},width=\textwidth]{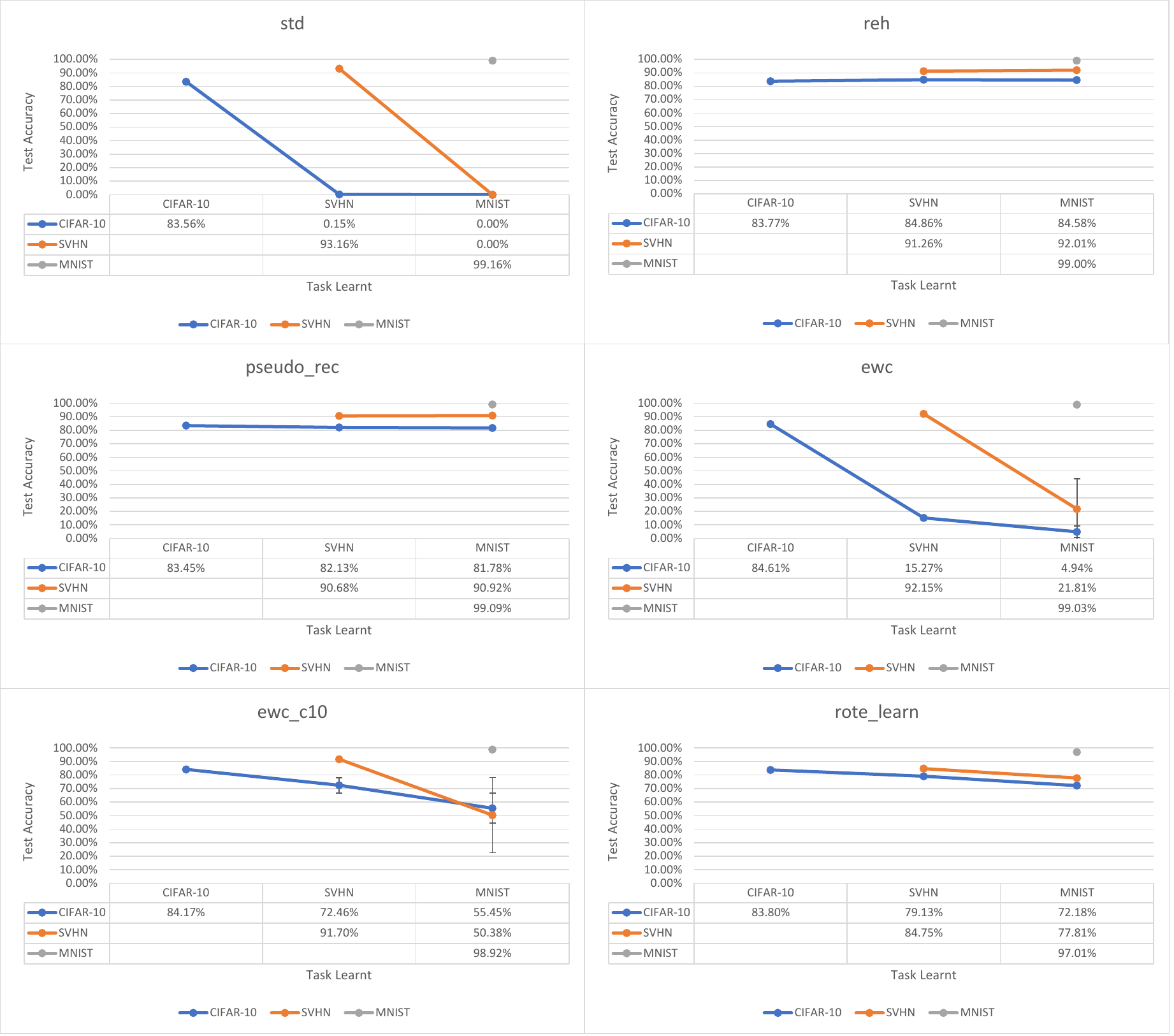}}
\caption{Average accuracy of the classification network for the $std$, $reh$, $pseudo\_rec$, $ewc$, $ewc\_c10$ and $rote\_learn$ conditions. The x-axis represents the task that has just been learnt and the lines represent the network's test accuracy on the various tasks trained so far. Error bars represent the standard deviation of each data point across the 3 trials. Non-visible error bars have smaller standard deviations than their data point.}
\label{results}
\end{center}
\vskip -0.2in
\end{figure*}

Although, the classification network we test is not the state of the art network for any of these datasets, it can still be trained to very respectable accuracy on all of the tasks (e.g. over 83\% on CIFAR-10) without using any special tricks. The results of all the experimental conditions are displayed in Figure~\ref{results}. The $std$ condition clearly shows CF because once a new task is learnt, the network does not correctly classify any of the previous tasks' images. The fact that the accuracy drops straight to 0\% on previous tasks seems dramatic, however it is very logical for a classification network that trains using cross-entropy because when training, the previous tasks' images do not appear at all and thus, the output neurons representing those classes quickly learn that they should never activate.

As expected, the $reh$ condition does not demonstrate CF, as the final task accuracies increased slightly from their initial values. This condition demonstrates that our network has the capacity to learn all three tasks to a high accuracy without needing any additional units.

The $pseudo\_rec$ condition also overcomes the CF problem as it too does not experience a dramatic drop in the previous tasks' accuracy when it learns a new task. In fact, it loses no more than 1.32\% accuracy every time a new task is presented and loses only 1.67\% of CIFAR-10 test accuracy after rehearsing both the other tasks. These differences in accuracy are absolute differences, which will remain consistent throughout this paper. For SVHN, pseudo-recursal resulted in a 0.24\% increase in accuracy after learning MNIST. This conveys that our network has the capability to retain almost all knowledge about the previous tasks without needing to store previous data, but rather by generating approximations of it as required.

The $ewc$ condition is barely resistant to the CF problem, managing to correctly classify 4.94\% and 21.81\% of the CIFAR-10 and SVHN datasets after all three tasks have been learnt. However, we hypothesised that this could be because each of the tasks' classes are represented by separate output units. Output units that represent the first task are never active for later tasks and thus, the pressure on these units to never activate on later tasks is likely greater than the pressure on these units to remember the previous task (from EWC). Therefore, in the $ewc\_c10$ condition we allowed the tasks to share their output units so that the units that are active for the first task are also active on subsequent tasks. We found that this lead to a dramatic improvement in EWC's ability to retain knowledge of previous tasks such that it could classify CIFAR-10 and SVHN to 55.45\% and 50.38\% accuracy after learning all tasks. This suggests that EWC is ineffective for learning tasks which do not share their output representations but is moderately effective when they do. We still find that pseudo-recursal clearly outperforms EWC as it loses only 1.67\% of CIFAR-10's accuracy compared to EWC's 28.72\%. Furthermore, it should be noted that over the 3 trials EWC's retention varied fairly dramatically, however pseudo-recursal consistently outperformed EWC across all trials by a large margin.

Compared to EWC and standard pseudo-rehearsal, the main disadvantage of this method is that a generative network is required for pseudo-rehearsal to work on our deep network. However, we also apply pseudo-rehearsal to the generative model so that the size of this network is constant. Another disadvantage of pseudo-recursal is that it takes considerably more training time because the generator must also be trained. Furthermore, the same number of pseudo-items as novel task's items are trained for both the classifier and the generator which results in twice as many mini-batches per epoch. However, we have not attempted to optimise the number of pseudo-items required and such a large increase may not be necessary.

The results for the $rote\_learn$ condition conveys that the classifier can retain the majority of its knowledge of past tasks, however pseudo-recursal still clearly outperforms it, retaining 9.6\% more accuracy on CIFAR-10 and 13.11\% more on SVHN. This demonstrates that using the GAN model is more effective than simply remembering past items.

\section{Discussion and Conclusion}
We have demonstrated that combining GANs with pseudo-rehearsal is an effective method for solving the CF problem. Pseudo-recursal has major advantages over other methods such as EWC because it does not require the network to grow for each new task and the network does not have any hard constraints on how individual neurons should learn the new task. In future research we aim to apply pseudo-recursal to a reinforcement learning task so that the network could learn to play novel Atari 2600 games, while retaining its ability to play previously trained games.

\section*{Acknowledgment}
We gratefully acknowledge the support of NVIDIA Corporation with the donation of the TITAN X GPU used for this research.

\bibliographystyle{IEEEtran}
\bibliography{IEEEabrv,mybibfile}

\begin{thebibliography}{10}
\providecommand{\url}[1]{#1}
\csname url@samestyle\endcsname
\providecommand{\newblock}{\relax}
\providecommand{\bibinfo}[2]{#2}
\providecommand{\BIBentrySTDinterwordspacing}{\spaceskip=0pt\relax}
\providecommand{\BIBentryALTinterwordstretchfactor}{4}
\providecommand{\BIBentryALTinterwordspacing}{\spaceskip=\fontdimen2\font plus
\BIBentryALTinterwordstretchfactor\fontdimen3\font minus
  \fontdimen4\font\relax}
\providecommand{\BIBforeignlanguage}[2]{{%
\expandafter\ifx\csname l@#1\endcsname\relax
\typeout{** WARNING: IEEEtran.bst: No hyphenation pattern has been}%
\typeout{** loaded for the language `#1'. Using the pattern for}%
\typeout{** the default language instead.}%
\else
\language=\csname l@#1\endcsname
\fi
#2}}
\providecommand{\BIBdecl}{\relax}
\BIBdecl

\bibitem{shin2017continual}
H.~Shin, J.~K. Lee, J.~Kim, and J.~Kim, ``Continual learning with deep
  generative replay,'' in \emph{Advances in Neural Information Processing
  Systems}, 2017, pp. 2994--3003.

\bibitem{mccloskey1989catastrophic}
M.~McCloskey and N.~J. Cohen, ``Catastrophic interference in connectionist
  networks: The sequential learning problem,'' \emph{Psychology of Learning and
  Motivation}, vol.~24, pp. 109--165, 1989.

\bibitem{abraham2005memory}
W.~C. Abraham and A.~Robins, ``Memory retention--the synaptic stability versus
  plasticity dilemma,'' \emph{Trends in Neurosciences}, vol.~28, no.~2, pp.
  73--78, 2005.

\bibitem{kirkpatrick2017overcoming}
J.~Kirkpatrick, R.~Pascanu, N.~Rabinowitz, J.~Veness, G.~Desjardins, A.~A.
  Rusu, K.~Milan, J.~Quan, T.~Ramalho, A.~Grabska-Barwinska \emph{et~al.},
  ``Overcoming catastrophic forgetting in neural networks,'' \emph{Proceedings
  of the National Academy of Sciences}, p. 201611835, 2017.

\bibitem{robins1995catastrophic}
A.~Robins, ``Catastrophic forgetting, rehearsal and pseudorehearsal,''
  \emph{Connection Science}, vol.~7, no.~2, pp. 123--146, 1995.

\bibitem{goodfellow2014generative}
I.~J. Goodfellow, J.~Pouget-Abadie, M.~Mirza, B.~Xu, D.~Warde-Farley, S.~Ozair,
  A.~Courville, and Y.~Bengio, ``Generative adversarial nets,'' in
  \emph{Advances in Neural Information Processing Systems}, 2014, pp.
  2672--2680.

\bibitem{radford2015unsupervised}
A.~Radford, L.~Metz, and S.~Chintala, ``Unsupervised representation learning
  with deep convolutional generative adversarial networks,'' \emph{arXiv
  preprint arXiv:1511.06434}, 2015.

\bibitem{robins1998local}
A.~Robins and M.~Frean, ``Local learning algorithms for sequential tasks in
  neural networks,'' \emph{Journal of Advanced Computational Intelligence and
  Intelligent Informatics}, vol.~2, no.~6, pp. 221--227, 1998.

\bibitem{mellado2017pseudorehearsal}
D.~Mellado, C.~Saavedra, S.~Chabert, and R.~Salas, ``Pseudorehearsal approach
  for incremental learning of deep convolutional neural networks,'' in
  \emph{Latin American Workshop on Computational Neuroscience}.\hskip 1em plus
  0.5em minus 0.4em\relax Springer, 2017, pp. 118--126.

\bibitem{kemker2017fearnet}
R.~Kemker and C.~Kanan, ``{F}ear{N}et: Brain-inspired model for incremental
  learning,'' \emph{arXiv preprint arXiv:1711.10563}, 2017.

\bibitem{draelos2016neurogenesis}
T.~J. Draelos, N.~E. Miner, C.~C. Lamb, J.~A. Cox, C.~M. Vineyard, K.~D.
  Carlson, W.~M. Severa, C.~D. James, and J.~B. Aimone, ``Neurogenesis deep
  learning,'' \emph{arXiv preprint arXiv:1612.03770}, 2016.

\bibitem{he2016deep}
K.~He, X.~Zhang, S.~Ren, and J.~Sun, ``Deep residual learning for image
  recognition,'' in \emph{Proceedings of Computer Vision and Pattern
  Recognition}, 2016, pp. 770--778.

\bibitem{kemker2017measuring}
R.~Kemker, M.~McClure, A.~Abitino, T.~Hayes, and C.~Kanan, ``Measuring
  catastrophic forgetting in neural networks,'' \emph{arXiv preprint
  arXiv:1708.02072}, 2017.

\bibitem{krizhevsky2012imagenet}
A.~Krizhevsky, I.~Sutskever, and G.~E. Hinton, ``Imagenet classification with
  deep convolutional neural networks,'' in \emph{Advances in Neural Information
  Processing Systems}, 2012, pp. 1097--1105.

\bibitem{salimans2016improved}
T.~Salimans, I.~Goodfellow, W.~Zaremba, V.~Cheung, A.~Radford, and X.~Chen,
  ``Improved techniques for training {GAN}s,'' in \emph{Advances in Neural
  Information Processing Systems}, 2016, pp. 2234--2242.

\end{thebibliography}

\end{document}